\documentclass[runningheads]{llncs}
\usepackage[T1]{fontenc}
\usepackage{graphicx}
\usepackage{booktabs}
\usepackage[misc]{ifsym}

\usepackage{mwe}

\usepackage{hyperref} 
\usepackage{xcolor}
\definecolor{darkgreen}{RGB}{81,164,82}
\definecolor{link_color}{RGB}{39,91,142}
\definecolor{myorange}{rgb}{1.0, 0.5, 0.0}

\hypersetup{
    colorlinks=true,       
    linkcolor=link_color,   
    filecolor=darkgreen,   
    urlcolor=link_color,    
    citecolor=darkgreen    
}

\usepackage{cite}

\usepackage{amssymb} 
\usepackage{bbm} 
\usepackage{graphicx} 
\usepackage{bm}

\usepackage{booktabs} 
\usepackage{multirow} 

\usepackage{listings}
\usepackage{xcolor}
\lstdefinestyle{pythonstyle}{
    language=Python,
    basicstyle=\fontsize{9pt}{9pt}\ttfamily,
    keywordstyle=\color{blue}\bfseries,
    commentstyle=\color{gray}\itshape,
    stringstyle=\color{red},
    numbers=left,
    numberstyle=\tiny\color{gray},
    frame=tb,
    columns=fullflexible,
    keepspaces=true,
    escapeinside={||},
    literate={'}{{\textquotesingle}}1, 
    morekeywords={cat, supcon_loss}    
}

\makeatletter
\newcommand*\bigcdot{\mathpalette\bigcdot@{.5}}
\newcommand*\bigcdot@[2]{\mathbin{\vcenter{\hbox{\scalebox{#2}{$\m@th#1\bullet$}}}}}
\makeatother
\newcommand{\lsupcon}{\mathcal{L}_{\text{SupCon}}}

\newcommand{\lssc}{\mathcal{L}_{\text{SSC}}}

\usepackage{float}
\floatstyle{plaintop}
\newfloat{codefloat}{htbp}{lop}
\floatname{codefloat}{Algorithm}

\usepackage{amsmath,amsfonts,bm}

\def\eqref#1{equation~\ref{#1}}

\def\1{\bm{1}}

\def\vlambda{{\bm{\lambda}}}

\def\vi{{\bm{i}}}

\def\vq{{\bm{q}}}

\def\vu{{\bm{u}}}

\def\vw{{\bm{w}}}
\def\vx{{\bm{x}}}
\def\vy{{\bm{y}}}
\def\vz{{\bm{z}}}

\def\evy{{y}}

\def\mU{{\bm{U}}}

\def\mW{{\bm{W}}}
\def\mX{{\bm{X}}}

\def\mZ{{\bm{Z}}}

\DeclareMathAlphabet{\mathsfit}{\encodingdefault}{\sfdefault}{m}{sl}
\SetMathAlphabet{\mathsfit}{bold}{\encodingdefault}{\sfdefault}{bx}{n}

\def\sR{{\mathbb{R}}}

\newcommand{\softmax}{\mathrm{softmax}}

\DeclareMathOperator*{\argmax}{arg\,max}

\begin{document}

\title{A Unified Contrastive Loss for Self-Training}


\author{Aurélien Gauffre \Letter   \and
Julien Horvat \and
Massih-Reza Amini}

\tocauthor{Aurélien Gauffre, Julien Horvat, Massih-Reza Amini}
\toctitle{A Unified Contrastive Loss for Self-Training}

\authorrunning{}
\authorrunning{A. Gauffre et al.}


\institute{Université Grenoble Alpes, CNRS/LIG, 150 place du Torrent, \\
38401 Saint-Martin d’Hères, France\\
\email{aurelien.gauffre@univ-grenoble-alpes.fr}}

\maketitle              

\begin{abstract}
Self-training methods have proven to be effective in exploiting abundant unlabeled data in semi-supervised learning, particularly when labeled data is scarce. While many of these approaches rely on a cross-entropy loss function (CE), recent advances have shown that the supervised contrastive loss function (SupCon) can be more effective. Additionally, unsupervised contrastive learning approaches have also been shown to capture high quality data representations in the unsupervised setting. To benefit from these advantages in a semi-supervised setting, we propose a general framework to enhance self-training methods, which replaces all instances of CE losses with a unique contrastive loss. By using class prototypes, which are a set of class-wise trainable parameters, we recover the probability distributions of the CE setting and show a theoretical equivalence with it. Our framework, when applied to popular self-training methods, results in significant performance improvements across three different datasets with a limited number of labeled data. Additionally, we demonstrate further improvements in convergence speed, transfer ability, and hyperparameter stability. The code is available at \url{https://github.com/AurelienGauffre/semisupcon/}.

\keywords{Semi-Supervised Learning   \and Contrastive Learning \and Classification \and Self-Training}
\end{abstract}
\section{Introduction}

Semi-supervised learning benefits significantly from advances in unsupervised representation learning, particularly through self-supervised approaches, which excel at efficiently extracting information from unlabeled data. Among these approaches, contrastive learning \cite{Hadsell2006DimensionalityMapping,Oord2018RepresentationCoding,He2020MomentumLearning,Chen2020ARepresentations} has been particularly effective in the field of computer vision. Moreover,  contrastive learning has been shown not limit its application to unsupervised settings. The standard practice for training deep neural networks in a supervised setting has traditionally involved using cross-entropy (CE) as the primary loss function. In recent works, \cite{Khosla2020SupervisedLearning} have developed a supervised contrastive loss function, dubbed \emph{SupCon}, which achieves highly discriminative representations and comparable or even superior results in accuracy. It uses information from the labels to create positive pairs, instead of relying on data augmentation to generate two different views of the same unlabeled sample. Specifically, positive instances (instances from the same class within a batch) are pushed closer together while pushing them away from negative instances (instances from other classes) in the embedding space. Recent research suggests that SupCon loss may have the potential to increase robustness and be less sensitive to various hyperparameter choices for data augmentation or optimizers \cite{Khosla2020SupervisedLearning, Gunel2021SupervisedFine-Tuning, Islam2021ALearning}.

However, while SupCon hinges on the presence of labeled training data, its unsupervised counterpart cannot leverage any label information. The primary objective of this study is to adapt the principles and advantages of SupCon to semi-supervised scenarios. Through the integration of both supervised and unsupervised aspects of contrastive learning, we introduce a Semi-Supervised Contrastive (SSC) framework which uses a single loss $\lssc$. Our approach enables the integration of existing  self-training techniques such as FixMatch \cite{Sohn2020FixMatch:Confidence}, allowing for a seamless transition between unsupervised and supervised paradigms.

Unlike CE loss trained with softmax activation function, contrastive loss does not provide directly a probability distribution needed to pseudo-label examples during self-training. To address this challenge, we propose a solution by introducing class prototypes and we establish a theoretical equivalence between classical cross-entropy and supervised contrastive learning with these prototypes. The main contributions of this work are threefold:

\begin{itemize}
\item We propose a new framework for semi-supervised learning based on a Semi-Supervised Contrastive loss $\mathcal{L}_{\text{SSC}}$ that handles labeled, pseudo-labeled and unconfident pseudo-labels examples at the same time.
 
 \item We show how to integrate class prototypes and establish a theoretical bridge between cross-entropy and supervised contrastive learning with prototypes.
 
 \item We apply our loss to FixMatch, a simple existing framework, and show significant improvement on three datasets and investigate the properties of our loss function, highlighting its faster convergence rate, adaptability to transfer learning and its stability to hyperparameters.

\end{itemize}
In the following, we begin by presenting the notations and background in Section \ref{sec:notations}. Next, in Section \ref{sec:method}, we introduce our proposed approach. Then, we discuss the experiments carried out on three benchmarks in Section \ref{sec:experiments}. Lastly, Section \ref{sec:conclusion} presents our conclusions.

\section{Notations and Background}
\label{sec:notations}
\subsubsection{Notations.} We will now introduce necessary notations and then show how they connect to previous related works. We use matrix notation rather than vectors, which provides a more convenient framework for presenting our approach. In the semi-supervised context, the batch is divided into a matrix $\mX$ consisting of $B$ labeled examples and their associated label vector $\vy^x$, and another matrix $\mU$ containing $\mu B$ unlabeled examples, where the integer $\mu$ denotes the factor size between $\mX$ and $\mU$. More specifically, we have :
\begin{align*}
  \mX &= \begin{bmatrix}
    \vx^\top_1 \\
    \vdots \\
    \vx^\top_B
  \end{bmatrix} \in {\mathcal{X}^B},
  \quad
    \vy^x = \begin{bmatrix}
  \evy_1 \\
  \vdots \\
  \evy_B 
  \end{bmatrix}\in {[1,...,K]^B},
  \quad
  \mU = \begin{bmatrix}
  \vu_1^\top \\
  \vdots \\
  \vu_{\mu B}^\top 
  \end{bmatrix} \in {\mathcal{U}^{\mu B}}
\end{align*}

We denote by $f: \mathcal{X} \cup \mathcal{U} \to \mathcal{Z} $, an encoder that maps examples into the hidden space $\mathcal{Z}\subseteq \mathbb{R}^h$. Then, a projection head $p$ maps embeddings into a probability distribution over the $K$ classes of our classification problem. In the context of self-training, \textit {pseudo-labels} refers to labels automatically assigned to unlabeled data by a model's highest confidence prediction, defined as the $\text{argmax}$ on the projection head $p$. The model is said to be confident in an unlabeled example if the maximum probability exceeds a threshold $\tau$.

Following most of the recent semi-supervised learning approaches based on consistency regularization, we employ both a weak data augmentation, denoted as $\alpha(.) $, and a strong augmentation, denoted as  $\mathcal{A}(.)$. During training, the encoder $f$ is trained to compute three distinct embeddings:

\begin{itemize}
  \item $ \mZ^x = f(\mX)=[\vz^x_1,\ldots,\vz^x_B]^\top \in \sR^{B \times d}$,  is the supervised embedding generated from the labeled training data. 
  \item $ \mZ^u = \begin{bmatrix}
        \mZ^{s1} \\
        \mZ^{s2}
      \end{bmatrix} = \begin{bmatrix}
        f(\mathcal{A}(\mU)) \\
        f(\mathcal{A}(\mU))
      \end{bmatrix} \in \sR^{2\mu B \times d} $  denote the embeddings produced by applying two stochastic strong augmentation to unlabeled data. Using two augmentations ensures at least one positive pair for each example in the batch.
  \item $ \mZ^{w} = f(\alpha(\mU))=[\vz^w_1,\ldots,\vz^w_{\mu B}]^\top \in \sR^{\mu B \times d} $ is the unsupervised embedding created through the application of weak data augmentation. This embedding is employed for the estimation of a confidence score and the generation of pseudo-labels in self-training approaches.
  
\end{itemize}
Finally, we define two set of labels associated to unlabeled examples :
\begin{itemize}
    \item $\vy^{u\uparrow} = \begin{bmatrix}
        \vq
       \\ \vq
      \end{bmatrix}$ where $\vq=\argmax p(\mZ^w)$ are the pseudo-labels computed with the weak-augmented examples. They are associated to unlabeled data with high confidence, above a given threshold $\tau$.
    \item $\vy^{u\downarrow} = \begin{bmatrix}
        \vi \\ 
        \vi
      \end{bmatrix}$ where $\vi = [1,2,...,\mu B]^\top $ are the labels associated to unlabeled examples $\tau$. This definition will ensures that these unconfident labels will only have a unique positive example associated with them.
\end{itemize}

We will briefly present the contrastive and semi-supervised learning losses of related work with the previous notation, and then show how our approach is connected to these losses functions.

\subsubsection{Supervised Contrastive Learning.} In \cite{Khosla2020SupervisedLearning}, labeled data is utilized to ensure that embeddings of samples with identical labels are pulled closer together, while ensuring that embeddings from samples with different labels are pushed farther apart. This is achieved by employing supervised embeddings $\mZ^x$ along with their corresponding labels $\vy^x$. 

The objective involves calculating, for each embedding $\mZ^x_i$ (referred to as the anchor), its cosine similarity with all other embeddings $\mZ^x_p$ that share the same label (referred to as positive pairs). Subsequently, this similarity is normalized by the sum of similarities across all pairs, following the principles of the classical InfoNCE loss \cite{Oord2018RepresentationCoding}. More precisely, we have:

\begin{equation}
\label{eq:supcon}
\lsupcon(\mZ^x,\vy^x) =  \frac{1}{|I|} \sum_{i \in I} \frac{-1}{|P(i)|} \sum_{p\in \mathcal{P}(i)} \log \Bigg( \frac{\exp({{\vz^x_i \bigcdot \vz^x_{p}}/T})}{\sum_{j\in I \setminus \{ i \}} \exp({{\vz^x_i \bigcdot \ \vz^x_{j}}/T})} \Bigg) 
\end{equation}

Where $I=\{1,...,B\}$ is the set of anchor indices, $P(i) = \{ p \in I \setminus \{ i \} : y^x_p = y^x_i \} $ is the set of positive examples associated with the example $i$ and $T$ is a temperature hyperparameter. Note that the labels  $\vy^x$ are used in the equation only to define of the positive pairs $\mathcal{P}$.

\subsubsection{Unsupervised Contrastive Learning.} As seen in methods SimCLR \cite{Chen2020ARepresentations}  or MoCo \cite{He2020MomentumLearning}, unsupervised contrastive learning relies on two strong augmentations for each instance within the unsupervised dataset $\mU$, and employs the InfoNCE loss \cite{Oord2018RepresentationCoding}. 

Unlike SupCon, which utilizes explicit labels to identify positive and negative samples, self-supervised losses operate under an unsupervised paradigm where labels are not provided. Consequently, in self-supervised learning, every augmented sample has only one positive pair, effectively constituting a specialized form of the SupCon loss. Based on the previous definition of $\vy^{u\downarrow}$, the self-supervised InfoNCE loss can be expressed simply as: 

\begin{equation}
\mathcal{L}_{\text{Self}}(\mZ^u) = \lsupcon(\mZ^u,\vy^{u\downarrow})
\label{L_self}
\end{equation}

Interpreting the unsupervised contrastive loss as a specific instance of SupCon is central to the design of our unified loss.

\subsubsection{Self-training \cite{AminiSelf-TrainingSurvey}.} Also commonly referred to as pseudo-labeling, self-training is a wrapper algorithm that is widely adopted in recent state-of-the-art semi-supervised learning approaches \cite{Chen2020BigLearners,Beyer2019S4L:Learning,Sohn2020FixMatch:Confidence,Berthelot2019ReMixMatch:Anchoring,Berthelot2019MixMatch:Learning,Li2021CoMatch:Regularization,Zheng2022SimMatch:Matching,Chen2023SoftMatch:Learning}. A classifier is first trained on the labeled training data, and then assigns iteratively pseudo-labels to unlabeled data and retrain the classifier with the augmented training set. Some approaches propose to use self-training in an online manner \cite{Berthelot2019MixMatch:Learning,Berthelot2019ReMixMatch:Anchoring,Sohn2020FixMatch:Confidence}. 

More specifically, FixMatch \cite{Sohn2020FixMatch:Confidence} apply a CE loss $\mathcal{L}_{x}$ to labeled examples $X$ whereas an extra unsupervised CE loss $\mathcal{L}_{u}$ is applied to unlabeled training examples with their associated pseudo-labels, only if the model confidence exceeds the threshold $\tau$ :
\begin{align}
\mathcal{L}_{x} &= \frac{1}{B}\sum_{i=1}^{B}H(p(\vz^x_i),y^x_i) \\
\mathcal{L}_{u} &= \frac{1}{\mu B} \sum_{i=1}^{\mu B}  H(p(\vz^u_i), y^{u\uparrow}_i) \mathbbm{1}(\max p(\vz^w_i) > \tau)
\end{align}

This unsupervised loss is based on consistency regularization principle, which enforces the model to become invariant to perturbations of the input, like strong augmentations. It has become central to many popular recent semi-supervised approaches in computer vision \cite{Laine2017TemporalLearning,Tarvainen2017MeanResults,Xie2020UnsupervisedTraining}.

Recently, adaptive thresholding strategies for generating pseudo-labels have been proposed in Dash \cite{Xu2021Dash:Thresholding}, FlexMatch \cite{Zhang2021FlexMatch:Labeling}, Adamatch \cite{Berthelot2022Adamatch:Adaptation}, and FreeMatch \cite{Wang2022FreeMatch:Learning}.
SoftMatch \cite{Chen2023SoftMatch:Learning} proposes to adjust pseudo-labels contributions based on their confidence levels by learning a parametric density function that adaptively assigns weights for each pseudo-labeled examples. 

On the other hand, CoMatch \cite{Li2021CoMatch:Regularization} and SimMatch \cite{Zheng2022SimMatch:Matching} introduce an additional contrastive loss that enforce similarity between representations having similar probability distribution. Other existing semi-supervised approaches \cite{Lee2022ContrastiveLearning,Wang2023DualMatch:Interaction} have already proposed to use the SupCon loss, as an extra regularization term applied to labeled or pseudo-labeled examples. Other than the self-training techniques we mentioned, very successful semi-supervised learning exist and often rely on self-supervised principles. This may involve using an additional regularization loss as in S4L \cite{Beyer2019S4L:Learning}, or using contrastive pre-training combined with distillation as in SimCLR V2 \cite{Chen2020BigLearners}, or using clustering approaches like PAWS \cite{Assran2022Semi-SupervisedSamples} or Suave and Daino \cite{Fini2023Semi-supervisedClustering}.

In contrast to all the aforementioned approaches, our method uses a single contrastive loss that handles both the labeled training data and all the unlabeled training examples at the same time, including those on which the model is not confident.

\section{Method}
\label{sec:method}
Our approach is a wrapper algorithm that can be easily adapted to various self-training algorithms. We will use FixMatch as an example to illustrate our approach because of its simplicity. However, our proposed approach is flexible enough to be applied to more complex self-training algorithms.
\subsection{Overview}

\begin{figure}[t!]
 \centering
 \hspace*{0cm}
 \includegraphics[scale=.7]{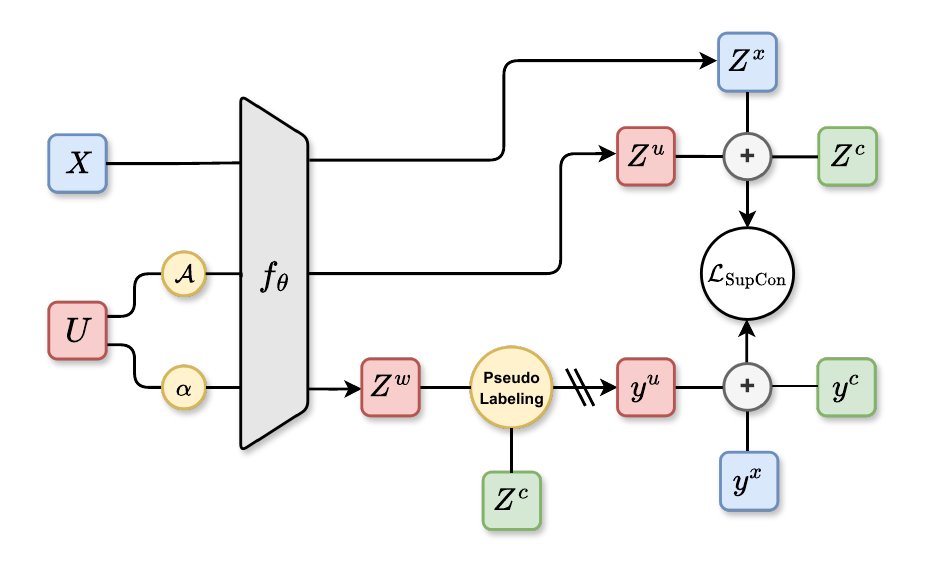} 
 \caption{\textbf{SSC framework}. $\mZ^x$, $\mZ^u$, and $\mZ^c$ are supervised, unsupervised, and prototype embeddings and $\vy^x$, $\vy^u$, and $\vy^c$ their corresponding labels, which  aim to define positive pairs in the loss. Both the triplets of embeddings and their corresponding labels are concatenated and subsequently input into the loss function. The weakly augmented embeddings $\mZ^w$ are used only during pseudo-labeling phase to compute $\vy^u$ and does not propagate gradient back. Strongly augmented embeddings $\mZ^u$ used two augmentations to ensure the existence of at least one positive pair for unconfident examples.}
\label{fig:SemiSupCon}
\end{figure}

In our approach, we aim to enhance the classical SupCon loss by integrating labeled, pseudo-labeled, and unlabeled examples on which the model is unconfident, simultaneously within the loss formulation. The fundamental architecture of our method is illustrated in Figure \ref{fig:SemiSupCon}.

Using the encoder $f$, we first compute $\mZ^x$, the embeddings of the labeled training data. In a similar way, we generate $\mZ^u$ by applying two strong data augmentations to the unlabeled training data, and $\mZ^w$ by applying a weak data augmentation.

\subsubsection{Unsupervised part $\mZ^u, \vy^u$.}
A pivotal innovation allowed in our framework is its way to handle all unlabeled examples in the loss, regardless of their confidence:
\begin{equation}
y^u_i=
\begin{cases}
y^{u\uparrow}_i & \text{if } \max p(\vz_i^w) > \tau \\
 y^{u\downarrow}_i+K & \text{otherwise}.
\end{cases}
\label{pl_eq}
\end{equation}

Concerning high confidence examples, we adopt a strategy similar to online self-training methods, like FixMatch, by using pseudo-labels previously defined as $\vy^{u\uparrow}$. However, rather than disregarding examples that have a posterior probability below the threshold $\tau$, we assign unique labels to them using $\vy^{u^\downarrow}$. Note that to make sure these labels are unique, values are shift by $K$ to not interfere with existing classes. 

This leads to the creation of singular positive pairs, mirroring the mechanics of unsupervised contrastive loss methods such as SimCLR as shown in \eqref{L_self}. By incorporating both confident and unconfident examples within $\vy^u$, our method is able to leverages all unlabeled training data. Note that, even if the loss does not directly depend on the weakly augmented embeddings, $\mZ^w$ is used to compute $\vy^u$.

\subsubsection{Centroid part $\mZ^c,\vy^c$.}
Computing labels for unlabeled examples $\vy^u$ requires a projection head $p$ that maps $\mZ^w$ into a distribution probability. However, training a model with a supervised contrastive loss does not produce directly such a classifier, which is why an extra training phase with cross-entropy is employed when doing classification with SupCon loss \cite{Khosla2020SupervisedLearning}. 

In order to address this issue, and to maintain a fully contrastive framework,  we propose the use of class prototypes \cite{Zhu2022BalancedRecognition,Cui2021ParametricLearning,Graf2021DissectingLearning}.
It consists in using $K$ trainable parametric centers $\mZ^c \in \mathbb{R}^{K \times d}$ that lie directly in the embeddings space. We define the label prototypes as $\vy^c = [1,2,...,K]^\top$ so that  the $k^{\text{th}}$ row of $\mathbf{Z}^c$ represents the prototype associated with class $k$. These parameters, initialized randomly, are then updated throughout the contrastive training  process similarly to all other embeddings. A novel aspect of our method is to use these prototypes to define a probability distribution  for a weakly augmented example $\mathbf{z}_i^w$, by applying a softmax function with temperature $T'$ to its cosine similarity with all prototypes:
\begin{equation}
    p(\vz_i^w) := \softmax (\frac{{\mZ^c \vz^w_i}}{T'})    
\end{equation}
Training with these prototypes  allows defining a classification head $p$, used to compute $\vy^u$, without the addition of an extra cross-entropy  loss. Further analysis and a connection with the cross entropy are discussed below.

\subsubsection{A unified loss $\mathcal{L}_{SSC}$.}
Finally, our loss, denoted as Semi-Supervised Contrastive (SSC) loss, can be easily expressed using SupCon and previously defined quantities:

\begin{equation}
\label{eq:SSC}
\lssc = \lsupcon\left( \begin{bmatrix}
       \mZ^x \\
       \mZ^u \\
       \mZ^c
      \end{bmatrix},\begin{bmatrix}
       \vy^x \\
       \vy^u \\
       \vy^c
      \end{bmatrix} \right)
\end{equation}

\begin{table}[b!]
\centering
\caption{Comparison of loss types used in various online self-training algorithms. The table indicates which type of loss, either CE (Cross-Entropy), Contrastive Learning (CL), or none ($\varnothing$) is applied to different parts of the input: $\mZ^x$ for embeddings of supervised examples, $\mZ^{u\uparrow}$ for high-confidence pseudo-labeled examples, and $\mZ^{u\downarrow}$ for unconfident examples (confidence less than threshold $\tau$).} 
\vspace{5pt}
\setlength{\tabcolsep}{10.5pt}
\begin{tabular}{lccc}
\toprule[.8pt] 
Example Types & $\mZ^x$ & $\mZ^{u\uparrow}$ & $\mZ^{u\downarrow}$ \\
\midrule
Standard Classification       & CE          & $\varnothing$      &  $\varnothing$      \\
Supervised Contrastive \cite{Khosla2020SupervisedLearning}               & CL & $\varnothing$      &  $\varnothing$      \\
Unsupervised Contrastive (e.g. \cite{Oord2018RepresentationCoding,Chen2020ARepresentations,He2020MomentumLearning})               & $\varnothing$ & $\varnothing$      &  CL        \\
\midrule
FixMatch, FlexMatch \cite{Sohn2020FixMatch:Confidence,Zhang2021FlexMatch:Labeling}               & CE          & CE                 &  $\varnothing$      \\
Fixmatch w. CR \cite{Lee2022ContrastiveLearning}  & CE          & CE                 &  CL       \\
CoMatch, SimMatch \cite{Li2021CoMatch:Regularization,Zheng2022SimMatch:Matching}      & CE          & CE + CL   &  $\varnothing$      \\

DualMatch \cite{Wang2023DualMatch:Interaction}      & CE + CL & CE + CL   &  $\varnothing$      \\

\midrule
Ours                   & CL & CL        &  CL        \\
\bottomrule[.8pt] 
\end{tabular}

\label{tab:recap}
\end{table}

Table \ref{tab:recap}, provides an overview of different state-of-the-art approaches that utilize labeled, pseudo-labeled and unconfident unlabeled data in their learning process. Comparatively, our proposed approach is the only one which takes advantage of all labeled and unlabeled training data for learning in a fully contrastive framework.

The pseudo-code of the proposed approach is provided in Algorithm 1. First,  the algorithm computes embeddings for labeled examples, $\mZ^x$,  unlabeled examples using a strong augmentation, $\mZ^u$, and for prototypes, $\mZ^c$. Labels associated with unlabeled examples $\vy^u$ are generated with our pseudo-labeling process using both $\mZ^c$ and the embeddings of weakly augmented examples $\mZ^w$ . Finally, it combines the labels of labeled examples, $\vy^x$, pseudo-labeled examples, $\vy^u$, and labels of the prototypes, $\vy^c$, to train the model using the SSC loss function, defined in Equation \ref{eq:SSC}.
\begin{codefloat}[t!]
\caption{Semi-Supervised Contrastive (SSC) Pseudocode}

\begin{lstlisting}[style=pythonstyle]
# Aug: strong augment, aug: weak augment
# T': softmax temperature for pseudo-labeling
# tau: threshold for pseudo-labeling
# lambda: weight
def training_step(X, U, y_x, prototypes):  
    # compute embeddings
    Z_x = f(X) 
    Z_u = cat(f(Aug(U)), f(Aug(U)))
    Z_w = f(aug(U)) 
    Z_c = prototypes
    # compute target using equation 5.
    y_u = compute_pseudo_labels(Z_w, Z_c, tau, T') 
    y_c = [|1|, ..., K]
    # compute loss 
    Z = cat(Z_x, Z_u, Z_c)
    y = cat(y_x, y_u, y_c)
    loss = supcon_loss(Z, y, |lambda|)
    return loss
\end{lstlisting}
\hrule
\end{codefloat}

\subsection{Weigthed Semi-SupCon Loss}

Similar to other mixed-loss frameworks that include parameters to balance between supervised, pseudo-labeled, or unsupervised parts, we extend the previously defined semi-supervised contrastive loss $\lssc$ to feature weights, by using an additional parameter $\lambda$:

\begin{equation}
\lssc(\mZ,\vy,\vlambda) =  \frac{1}{\sum_{k \in I} \lambda_k}  \sum_{i \in I} \frac{-\lambda_i}{|P(i)|} \sum_{p\in \mathcal{P}(i)} \log \Bigg( \frac{\exp({{\vz_i \bigcdot \vz_{p}}/T})}{\sum_{j\in I \setminus \{ i \}} \exp({{\vz_i \bigcdot \ \vz_{j}}/T})} \Bigg) 
\label{eq:ssc_weight}
\end{equation}

These weights provide a mechanism to give higher importance to some anchors. In practise, we use a very simple strategy where we use a constant value that depends only on the nature of the anchor:
\begin{equation}
\lambda_i = 
\begin{cases} 
\lambda^x & \text{if } \vz_i \in {\mZ}^x\\
\lambda^{u\uparrow} & \text{if } \vz_i \in {\mZ}^{u\uparrow}\\
\lambda^{u\downarrow} & \text{if } \vz_i \in {\mZ}^{u\downarrow}\\
\lambda^c & \text{if } \vz_i \in {\mZ}^c\\
\end{cases}
\end{equation}
More advanced approaches using adaptive weighting can be easily implemented, for instance with weights based on the confidence of the classifier $p$ \cite{Li2021CoMatch:Regularization,Chen2023SoftMatch:Learning}. From now, $\lssc$ always refer to this weighted version of the loss.

\subsection{Link with cross-entropy}

We now establish a relationship between cross-entropy (CE) loss and  our framework using contrastive learning loss using prototypes, in the classical supervised framework, under mild assumptions. As already observed in previous work \cite{Sohn2016ImprovedObjective}, both loss functions have inherent similarities, particularly in treating negative embeddings similarly to the weights of a linear classification layer. Our prototype-based approach builds on this analogy. If we remove the bias of the last projection layer, the CE loss $H$ can be expressed in terms of the weights of the final linear projection layer $\mW \in \mathbb{R}^{K \times d}$ as such :

\begin{align}
H(\mZ^x,\vy) &= \frac{1}{B} \sum_{i=1}^{B} -\log\softmax (\mW \vz^x_i)_{y_i} \notag \\
&= \frac{1}{B} \sum_{i=1}^{B} -\log \frac{\exp({{\vz^x_i \cdot \vw_{y_i}}})}{\sum_{k=1}^K \exp({{\vz^x_i \cdot \vw_{k}}})}
\end{align}

If we set the temperature of the SupCon loss to $T=1$, and ensure the normalization of all embeddings, it is now easy to see that by replacing the weights $\mW$ of the last layer with the prototypes $\mZ_c$, we get :
\begin{equation}
\label{eq:H}
H(\mZ^x,\vy) = \frac{1}{B} \sum_{i=1}^{B} \lsupcon \left( \begin{bmatrix} \vz^x_i \\ \mZ^c \end{bmatrix}, \begin{bmatrix} y_i \\ \vy^c \end{bmatrix} \right) 
\end{equation}
The CE loss is equivalent to applying separate SupCon losses to each example, each of which has only one positive pair that is its class prototype. Both losses aim fundamentally to learn prototypes $\mZ^c$ or equivalently weights $\mW$ to be aligned with their corresponding feature vectors given by the labels, which supports the use of the prototypes to learn a similar distribution probability $p$ on the embeddings space.

\section{Experiments}
\label{sec:experiments}

In the following section, we present our experimental setup, compare our method with established self-training approaches, and evaluate the impact of individual components through an ablation study. We also investigate the transfer performance of our approach and its synergy with self-supervised pre-training, focusing on convergence speed. Finally, we analyze the stability of the hyperparameters in our proposed loss function.

\subsection{Experimental setup}
\label{sec:ES}
Our framework is evaluated on three classical benchmark datasets: CIFAR-100 \cite{AlexKrizhevsky2009LearningImages}, STL-10 \cite{Coates2011AnLearning}, and SVHN \cite{Netzer2011ReadingLearning}. For each dataset, we explore two splits, keeping a limited number of 4 and 25 labeled examples per class. We conducted each experiment using 3 random seeds and present both the mean and the standard deviation for each experiment. Following the setup in \cite{Sohn2020FixMatch:Confidence}, baseline models are reported for 1024 epochs, where an epoch is arbitrary defined as $2^{10}$ steps following the literature. However, to demonstrate the efficiency of our approach, we  only train with $\lssc$ on 256 epochs.

We use a Wide ResNet WRN-28-2 \cite{Zagoruyko2016WideNetworks} for all experiments on CIFAR-100 and SVHN, while a larger WRN-37-2 is used for STL-10. Additionally, on top of these architectures, we added a projection head as mentioned in SupCon, which consists of a 2-layer MLP with dimensions of 128 for WRN-28-2 and 256 for WRN-37-2 (following the dimension of the original projection used with CE). For FixMatch, the strong augmentation used is RandAugment \cite{Cubuk2020Randaugment:Space}. 

It is important to note that although RandAugment is commonly used in semi-supervised settings, it is not specifically designed for contrastive learning. Nevertheless, we decided to keep the same augmentation parameters as those used in FixMatch. To be fair, we adopted the exact hyperparameters from the original work, including all optimizer settings such as learning rate, schedule, weight decay, batch size $B$ and ratio $\mu$. Concerning the extra hyperparameter introduced in our framework, we keep them the same for all the experiments. We take $T=0.01$ which is a common temperature value used in SupCon loss, and we set $T'=0.04$. 

Tuning this last parameters is actually equivalent to tuning the pseudo-labeling threshold $\tau$, which is kept at $\tau=.95$ to be consistent with Fixmatch. Indeed, increasing $T'$ will cause the posterior distribution $p$ to approach the uniform distribution, which will have the same effect on pseudo-labeling as increasing $\tau$. We chose to give the same importance to all embeddings by setting $\lambda^x=\lambda^{u\uparrow}=\lambda^c=1$ except for unconfident one where we set $\lambda^{u\downarrow}=0.2$

\subsection{Experimental Results}
\label{sec:ER}

\subsubsection{Performance of $\lssc$}
\label{sec:semisupcon}
We begin our evaluation by comparing FixMatch with and without the use of our proposed semi-supervised contrastive loss against other leading self-training approaches. We conduct this comparison on CIFAR-100 and SVHN datasets, employing both 4 and 25 labeled training samples. We report the results of the state-of-art approaches that have been previously found in the literature\footnote{\url{https://github.com/microsoft/Semi-supervised-learning/blob/main/results/classic_cv.csv}} and in order to see the effect of the proposed approach, we ran FixMatch with and without SemiSupCon loss on our servers.

\begin{table}[htbp]
\centering
\caption{Top-1 validation accuracy (\%) of various self-training methods compared to FixMatch, without and with the integration into our proposed wrapper approach (denoted as FixMatch w. $\mathcal{L}_{SSC}$) obtained after convergence.}
\vspace{5pt}
\setlength{\tabcolsep}{1.6pt} 
\begin{tabular}{lcccccccc}
\toprule[.7pt]
Dataset & ~~&\multicolumn{3}{c}{CIFAR-100} & & \multicolumn{3}{c}{SVHN} \\
\cline{1-1} \cline{3-5}\cline{7-9} 
labels/class & ~~  & 4  & & 25 & ~~ & 4  &  & 25 \\
\cline{1-1} \cline{3-3} \cline{5-5}\cline{7-7} \cline{9-9}
$\Pi$-Model \cite{Laine2017TemporalLearning}                  &   ~~       & 12.87 $\pm$\scriptsize{1.25} & &39.92 $\pm$\scriptsize{0.61}&  ~~& 22.62 $\pm$\scriptsize{5.36} & &86.45 $\pm$\scriptsize{0.42}\\

MixMatch \cite{Berthelot2019MixMatch:Learning}                  &   ~~       & 20.05 $\pm$\scriptsize{0.29} & &50.42 $\pm$\scriptsize{0.62}&  ~~& 20.37 $\pm$\scriptsize{5.78} & &96.29 $\pm$\scriptsize{0.2} \\
VAT \cite{Miyato2019VirtualLearning}                  &   ~~       & 23.58 $\pm$\scriptsize{2.57} & &46.83 $\pm$\scriptsize{0.57}&  ~~& 23.01 $\pm$\scriptsize{6.59} & &95.41 $\pm$\scriptsize{0.13} \\
RemixMatch \cite{Berthelot2019ReMixMatch:Anchoring}                  &   ~~       & 42.91 $\pm$\scriptsize{0.01} & &65.23 $\pm$\scriptsize{0.32}&  ~~& 68.73 $\pm$\scriptsize{18.79} & & 93.62 $\pm$\scriptsize{1.09} \\
UDA  \cite{Xie2020UnsupervisedTraining}                 &   ~~       & 46.56 $\pm$\scriptsize{2.06} & &65.63 $\pm$\scriptsize{0.28}&  ~~& 97.71 $\pm$\scriptsize{0.02} & &97.72 $\pm$\scriptsize{0.03} \\
\midrule
FixMatch                   &   ~~       & 46.62 $\pm$\scriptsize{2.38} & &65.35 $\pm$\scriptsize{0.62}&  ~~& 97.83 $\pm$\scriptsize{0.03} & &97.96 $\pm$\scriptsize{0.07} \\
FixMatch w. $\lssc$   &  ~~ & \textbf{48.45 $\pm$\scriptsize{1.32}} & &\textbf{67.05 $\pm$\scriptsize{0.48}} &  & \textbf{97.94 $\pm$\scriptsize{0.06}} & &\textbf{98.15 $\pm$\scriptsize{0.08}} \\
\midrule
Fully supervised (CE) &  & 77.45 $\pm$ \scriptsize{0.02} & & 77.54 $\pm$ \scriptsize{0.23} &  & 97.91 $\pm$ \scriptsize{0.02} & & 97.91 $\pm$ \scriptsize{0.01} \\

\bottomrule[.7pt]
\end{tabular}
\label{tab:comp}\end{table}

Based on the results presented in Table \ref{tab:comp}, it comes that UDA \cite{Xie2020UnsupervisedTraining} demonstrates comparable performance to FixMatch. However, when employing FixMatch with the proposed approach, denoted as $\mathcal{L}_{SSC}$, the method notably enhances its competitiveness, particularly evident when training the model with only $4$ labeled examples per class. These results underscore the effectiveness of our approach in leveraging all unlabeled data, particularly in scenarios where labeled data is scarce.

\subsubsection{Transfer Performance}
Classical semi-supervised learning benchmarks typically require training models from scratch, a process that consumes considerable time. Due to these constraints, certain studies advocate for leveraging pre-trained models in semi-supervised approaches \cite{Wang2022USB:Classification,Fini2023Semi-supervisedClustering}.

In this line, we explore the efficacy of integrating self-supervised pre-training using MoCo v2 \cite{Chen2020ImprovedLearning}, into our methodology. Specifically, in this section, we use a ResNet-50 architecture\footnote{We follow a standard adaptation of ResNet for smaller images, replacing the initial 7x7 convolutional layer with a 3x3 kernel and removing the final max pooling layer.} either trained from scratch or starting with MoCo v2 weights obtained after pretraining on ImageNet on 800 epochs\footnote{\url{https://github.com/facebookresearch/moco}}.

\begin{figure}[t!]
 \centering
 \hspace*{0cm}
 \includegraphics[scale=.35]{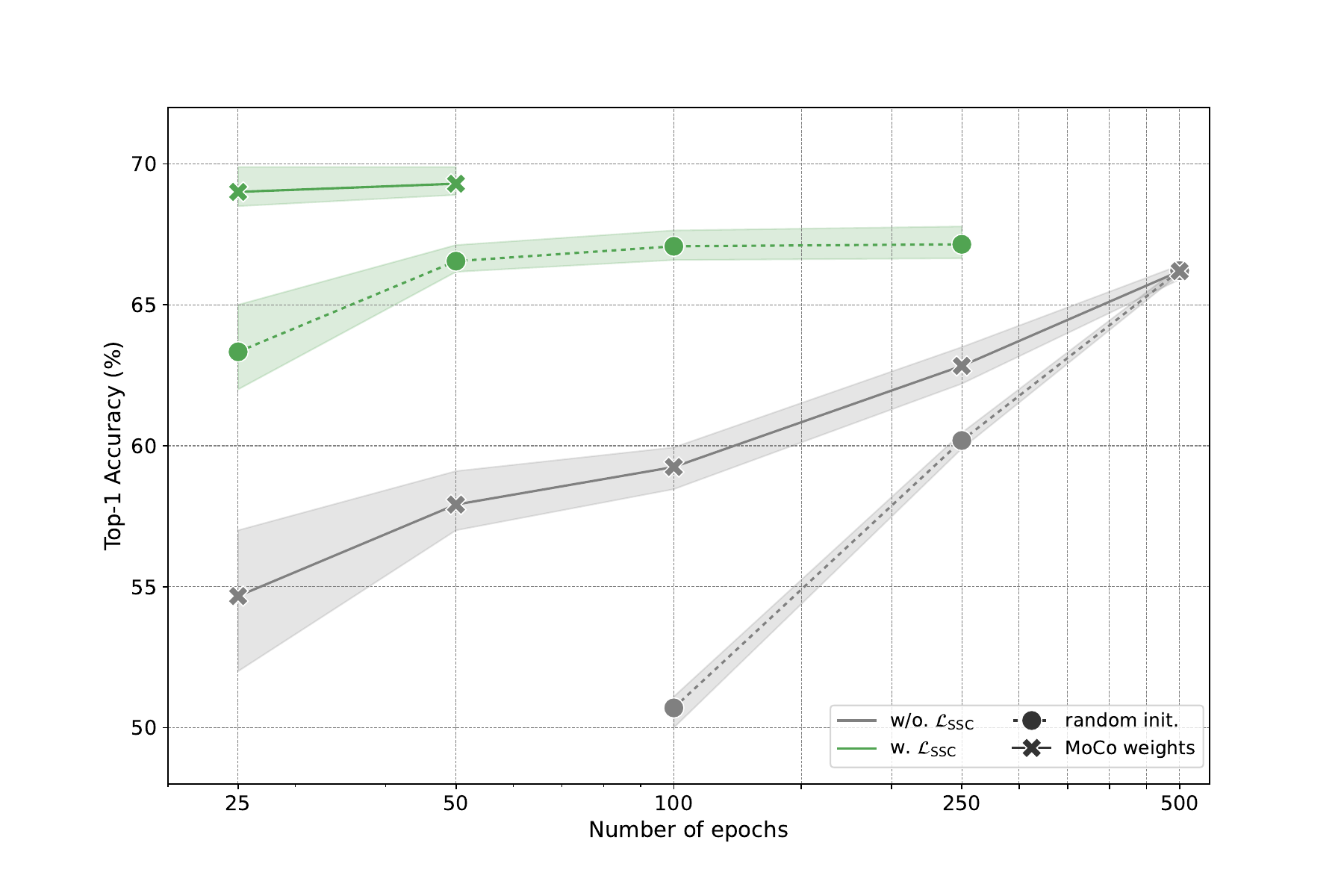}
 \caption{Transfer Performance with FixMatch on CIFAR-100 with 25 labels per class. The color gray (resp. green) corresponds to FixMatch without $\mathcal{L}_{SSC}$ (resp. with $\mathcal{L}_{SSC}$), while the dashed (resp. solid) line represents training from scratch (resp. using MoCo v2 weights). }
\label{fig:G}
\end{figure}

Figure \ref{fig:G} plots the Top-1 accuracy in percentage with respect to the number of epochs. We first observe that, in addition to having higher accuracy, using $\lssc$ loss requires substantially fewer epochs to converge. With only 50 epochs, training with $\lssc$ from scratch already outperforms the standard approach with 500 epochs. Only 25 epochs are needed when using pretrained weights, achieving a significantly higher validation accuracy of 69.3\%. Using all unlabeled data, including instances where the model has lower confidence, facilitates efficient training.

As noted, we observe a significant gain from the self-supervised pre-training with $\lssc$, which is not the case when using the classical CE loss. This underscores that our proposed loss seem to facilitate a smoother transition from pre-training methods, particularly those with a contrastive nature like MoCo.
\label{sec:TP}

\subsubsection{Ablation study}
\label{sec:AS}
\begin{table}[t]
\centering
\caption{Ablation study on CIFAR-100 and STL-10 with 256 epochs for all experiments. Starting from FixMatch (1), we use a double strong augmentation (2) and add a separate SimCLR loss on all unlabeled data (3). On the other hand, we try to remove the unconfident embeddings from Fixmatch with $\lssc$ (4), and then to use these unconfident examples in a separate SimCLR loss. (5). }
\vspace{5pt}
\setlength{\tabcolsep}{2.6pt}
\begin{tabular}{lcccc}
\toprule[.7pt]
Dataset & \multicolumn{2}{c}{CIFAR-100} & \multicolumn{2}{c}{STL-10} \\
\cmidrule(lr){1-1} \cmidrule(lr){2-3} \cmidrule(lr){4-5}
labels/class       & 4  & 25 & 4 & 25 \\
\cmidrule(lr){1-1} \cmidrule(lr){2-3} \cmidrule(lr){4-5}
(1) Base (FixMatch)                                  & 43.96 $\pm$\scriptsize{1.58} & 64.13 $\pm$\scriptsize{0.23} & 63.13 $\pm$\scriptsize{6.48} & 88.57 $\pm$\scriptsize{1.19} \\
(2) w.  Double Aug.                                  & 44.57 $\pm$\scriptsize{1.14} & 65.98 $\pm$\scriptsize{0.31} & 66.14 $\pm$\scriptsize{5.82} & 88.71 $\pm$\scriptsize{1.26} \\
(3) w. Double Aug. + $\mathcal{L}_{\text{Self}}$   & 42.16 $\pm$\scriptsize{1.76} & 64.47 $\pm$\scriptsize{0.18} & 55.74 $\pm$\scriptsize{5.33} & 86.22 $\pm$\scriptsize{0.96} \\
\midrule
(4) $\lssc$, $\lambda_{u\downarrow}=0$             & 45.03 $\pm$\scriptsize{1.26} & 65.96 $\pm$\scriptsize{0.14} & 70.21 $\pm$\scriptsize{6.95} & 87.69 $\pm$\scriptsize{1.65} \\
(5) $\lssc$, $\lambda_{u\downarrow}=0 + \mathcal{L}_{\text{Self}}$ & 43.06 $\pm$\scriptsize{1.42} & 62.12 $\pm$\scriptsize{0.17} & 56.45 $\pm$\scriptsize{7.21} & 86.72 $\pm$\scriptsize{1.36} \\
 (6) $\lssc$ & $\mathbf{46.53 \pm 1.18}$ & $\mathbf{66.28 \pm 0.22}$ & $\mathbf{73.21 \pm 6.73}$ & $\mathbf{88.94 \pm 1.26}$ \\
\bottomrule[.7pt]
\end{tabular}

\label{tab:E}
\end{table}

In order to investigate the effect of different components of $\lssc$, we perform an extensive ablation study, as reported in table \ref{tab:E} on CIFAR-100 and STL-10 by training the models with 256 epochs. 
We observed that using two strong augmentations slightly enhances the FixMatch technique. However, adding the self-supervised SimCLR loss tends to degrade performance, as already observed in \cite{Lee2022ContrastiveLearning}. Similarly, ignoring unconfident embeddings $\mZ^{u\downarrow}$ or applying them with a separate SimCLR loss also degrades the performance of $\lssc$.  The use of $\lssc$ consistently achieves the highest accuracy. These results justify our decision to incorporate them directly into our loss, thus facilitating global interaction with all other embeddings and prototypes.
\subsubsection{Hyperparameter stability analysis}
\label{sec:hyper}
\begin{figure}[H]
 \centering
 \includegraphics[scale=.4]{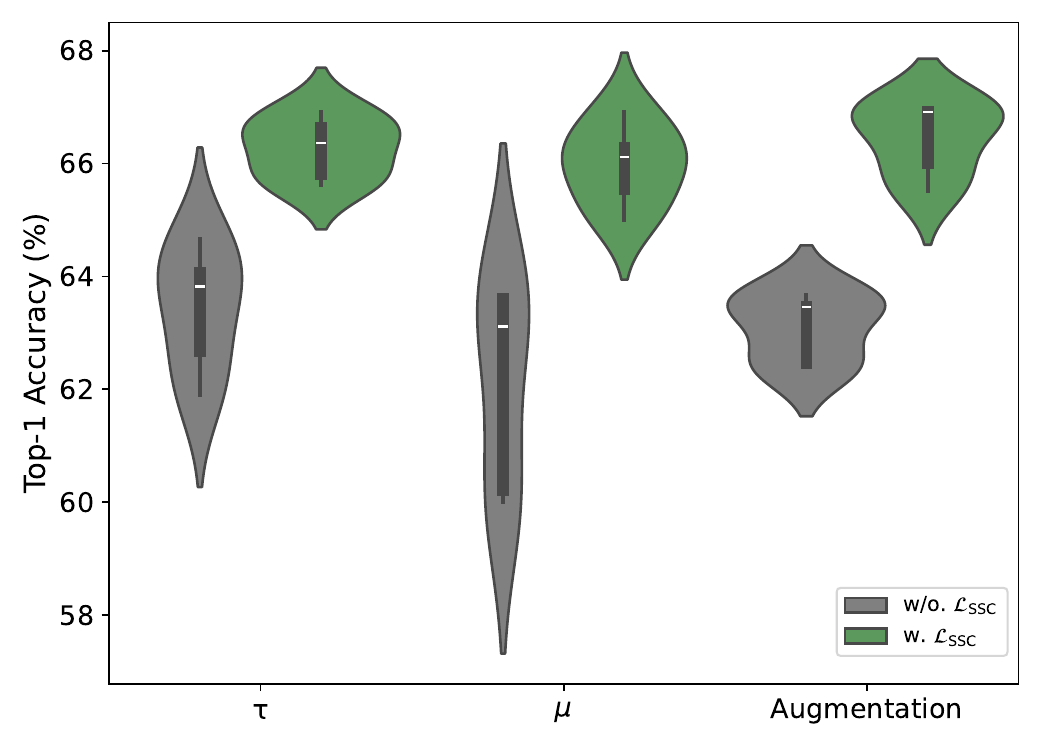}
 \caption{Hyperparameter stability analysis with FixMatch on CIFAR-100. For each hyperparameter, 10 experiments are conducted using the same seed with different values uniformly distributed in the followings range : $\tau\in [.9,0.98]$, $\mu \in \{3,...,12\}$ and RandAugment strength parameter in \{3,...,20\}. All experiments are run on 256 epochs with 25 labels/class.}
\label{fig:I}
\end{figure}

We examine the sensitivity of our framework to classical self-training hyperparameters, such as the pseudo-labeling confidence threshold $\tau$, the imbalance ratio between labeled and unlabeled examples in the batch $\mu$, and the strength of strong augmentation $\mathcal{A}(\cdot)$. Figure \ref{fig:I} illustrates the distribution of model performances across various hyperparameter settings.

Our contrastive approach, depicted in green, demonstrates significantly lower variance concerning the $\tau$ and $\mu$ parameters compared to alternative methods. However, both approaches appear equally sensitive to the augmentation strength. This outcome was anticipated since our contrastive framework relies on $\mathcal{A}(\cdot)$ for both consistency regularization and unsupervised contrastive learning through the utilization of unconfident embeddings $\mZ^{u\downarrow}$.

\section{Conclusion}
\label{sec:conclusion}
In this paper, we introduce a new semi-supervised contrastive framework that combines SupCon with an unsupervised contrastive loss, effectively operating within a self-training setting. The proposed framework allows taking advantage of labeled, pseudo-labeled, and unconfident examples simultaneously in the training process. 

Moreover, we propose the incorporation of class prototypes into contrastive learning to derive class probabilities, enhancing the interpretability and performance of the model.
By applying our approach to the FixMatch framework, we observe substantial performance gains across three datasets. Our method exhibits rapid convergence, benefits from pretraining, and showcases stability across various hyperparameters, underscoring its effectiveness and reliability in semi-supervised learning scenarios.

Future research avenues may explore further enhancements to the contrastive learning framework, such as incorporating domain-specific knowledge or adapting the framework to handle noisy or incomplete data. Additionally, investigating the interplay between contrastive learning and other semi-supervised learning techniques could lead to synergistic approaches with even greater performance gains.

\section*{Acknowledgement}
This project was provided with computer and storage resources by GENCI at
IDRIS thanks to the grant 2024-AD011014050R1 on the supercomputer Jean Zay on V100 and A100
partitions.

\bibliographystyle{splncs04}

\end{document}